\documentclass[11pt,a4paper]{article}
\usepackage[hyperref]{acl2021}
\usepackage{times}
\usepackage{latexsym}
\usepackage{subfigure}
\usepackage{graphicx}

\usepackage{microtype}

\aclfinalcopy 

\title{NEUer at SemEval-2021 Task 4: Complete Summary Representation by Filling Answers into Question for Matching Reading Comprehension}

\author{Zhixiang Chen\textsuperscript{1$\dagger$}, Yikun Lei\textsuperscript{1$\dagger$}, Pai Liu\textsuperscript{1,2$\dagger$}, Guibing Guo\textsuperscript{1*}\\
	$^1$Northeastern University, Shenyang, China \\
	$^2$Westlake University, Hangzhou, China
	\\
	\textsuperscript{$\dagger$} These three authors contributed equally.\\
	\tt \{zxchen8830, ethanlei9931, pailiu1998\}@gmail.com \\
   \tt guogb@swc.neu.edu.cn \\
   
}

\date{}

\begin{document}
\maketitle

\begin{abstract}
	SemEval task 4 aims to find a proper option from multiple candidates to resolve the task of machine reading comprehension. Most existing approaches propose to concat question and option together to form a context-aware model. However, we argue that straightforward concatenation can only provide a coarse-grained context for the MRC task, ignoring the specific positions of the option relative to the question. In this paper, we propose a novel MRC model by filling options into the question to produce a fine-grained context (defined as summary) which can better reveal the relationship between option and question. We conduct a series of experiments on the given dataset, and the results show that our approach outperforms other counterparts to a large extent.
\end{abstract}

\section{Introduction}
In order to make the computer understand, represent and express better, we study the ability of MRC
to understand Abstract definitions\citep{spreen1966parameters,changizi2008economically} in the Reading Comprehension
of Abstract Meaning task(ReCAM). In ReCAM task, there are two kinds of abstract definitions, one
is imperceptibility and the other is nonspecificity. To evaluate the model performance comprehensively,
three subtasks are designed\citep{zhengsemeval}:

Subtask 1) ReCAM-Imperceptibility  -  A  passage, a question with a placeholder, and answers used to fill in the placeholder are given.  The answers are all imperceptibility words so as to evaluatethe model’s ability to comprehend these imperceptibility words.

Subtask 2) ReCAM-Nonspecificity  -  A  passage, a question with a placeholder, and answers used to fill in the placeholder are given.  The answers are all Nonspecificity words so as to evaluatethe model’s ability to comprehend these Nonspecificity words.

Subtask 3) To provide more insights into the relationship between two abstract views, it requires to test the performance of the model trained on one data set and evaluated on the other one.

\begin{figure*}[t!]
	\centering
	\includegraphics[width=15cm]{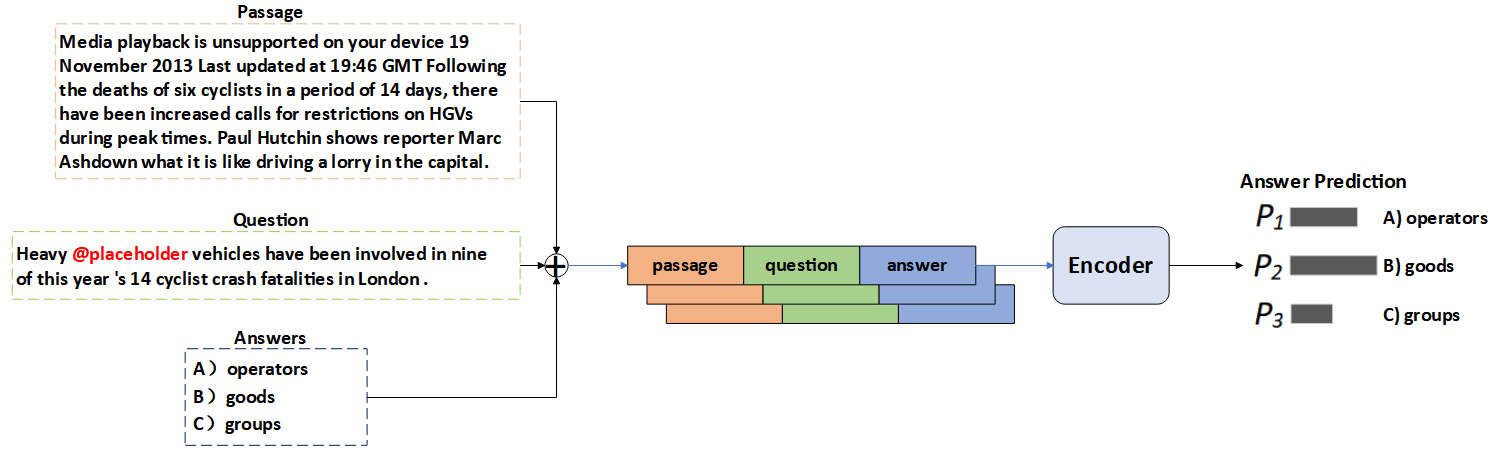}
	\caption{The framework of previous MCQA model. The abstract word in answer has a polysemy phenomenon. Entering question and answer separately cannot eliminate ambiguity.\label{fig1} }
\end{figure*}

Because question is obtained by hollowing out summary of passage and answer is abstract words to
fill in\citep{zhengsemeval}, there is a semantically complementary relationship between the question and answer. As shown in the Figure~\ref{fig1}, in the previous work\citep{jin2020mmm,zhu2020dual}, the model inputs passage, question and answer independently. It can only provide a coarse-grained context for the MRC task, ignoring the specific positions of the option relative to the question.  Quesiton cannot be a complete sentence, and the answer lacks
contextual information about the question. To solve this problem, we introduce {\bf Com}plete {\bf S}ummary {\bf R}epresentation by
Filling Answers into Question for Maching Reading Comprehension(ComSR). ComSR convert the original question and answer
into summary. In this way, we can turn the problem into matching correct summary for the passage. By replacing the answer option into the question placeholder, the original question has higher semantic integrity, and
the original answer obtains the context information from the question to eliminate language ambiguity such as polysemy.

Based on the results, ComSR is superior to pre-trained language models and the latest MRQA models,
with the accuracy rate of 64.76\% in subtask 1, 64.86\% in subtask 2, and shows great generalization ability
in Subtask 3.
\begin{figure*}[t!]
	\centering
	\includegraphics[width=15cm]{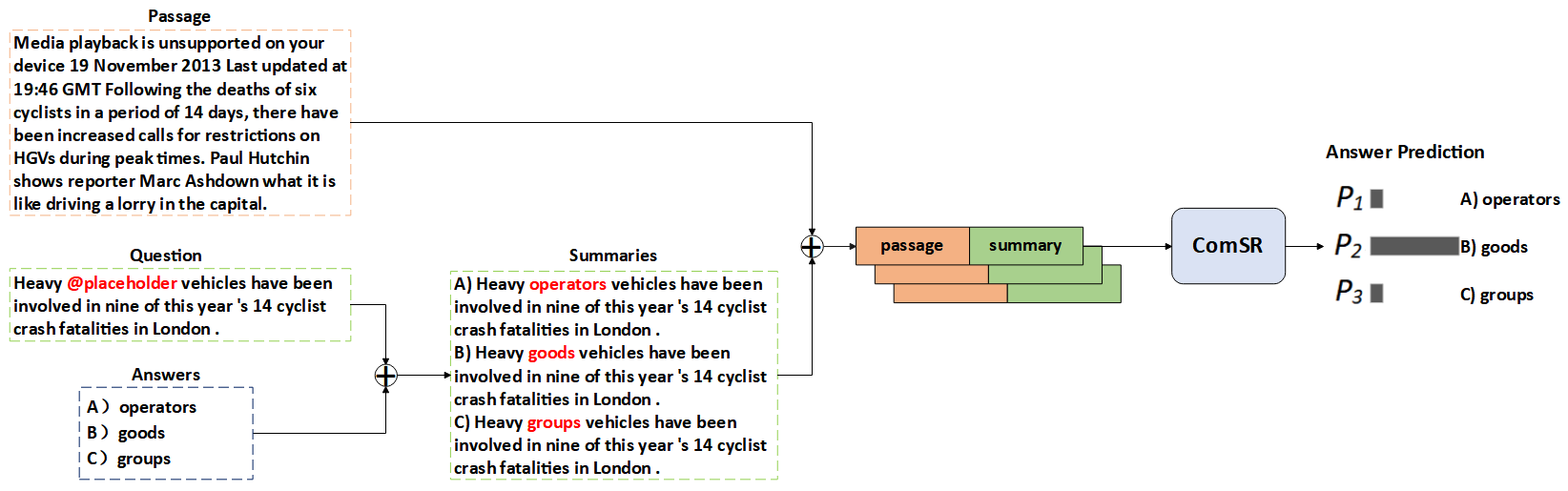}
	\caption{The framework of our model. By replacing the answer option into the question placeholder, answer obtains the context information from question, so the ambiguity is eliminated. \label{fig2} }
\end{figure*}

\section{System overview}
Figure~\ref{fig2} illustrates the overall structure of ComSR. Since just one summary option belongs to the passage, we input passage and summary together to better reflect the corresponding relationship
between the passage and the summaries. As BERT has been shown to be a powerful feature extractor for
various tasks\citep{ran2019option}, we use BERT as the encoder to extract the vector representation of the semantic meaning of the passage and the
summaries, as well as the relationship between the passage and the summaries. Finally, we use a classifier
layer to convert the hidden state into the final selection of the model.

\subsection{Input Layer}
The initial question uses "@placeholder" to represent the abstract word, which makes the question
sentence not a complete sentence. We replace the "@placeholder" in question with the answer word to get
the summary $S$. As a complete sentence, $S$ provides rich context information to answer, thus effectively
eliminating the ambiguity caused by polysemy in answer. Then the passage $P$ and the summary $S$ are
concatenated, using [CLS] at the beginning and [SEP] at the end of sentences to get input vector.
\begin{equation}\label{eq1}
	{x}=[{p} ; {s}]
\end{equation}
where $x$ denotes input vector, $p$ denotes passage vector and $s$ denotes summary vector.

\subsection{Encoder}
Although Pre-trained language models have achieved good results in various NLP tasks, most existing
MCQA datasets are small in size, pre-trained language models are hard to get adequate training. To solve this problem, we use the encoder obtained through multi-stage multi-task learning
in MMM \citep{jin2020mmm} as the initial encoder\footnote{\url{https://drive.google.com/drive/folders/1EECS9na9PpX9CO_cCzYj9FDkiBvOpyxv}}. The reasons are as follows. First, the
use of multiple data sets for training makes up for the shortcomings of the small amount of data in a
single data set. It uses four similar task data sets, MultiNLI\citep{williams2017broad},
SNLI \citep{young2014image}, DREAM \citep{sun2019dream}, and RACE \citep{lai2017race} for training, which
has significantly more data than a single MCQA data set, thus solving the defect of insufficient data
in a single data set. Second, the model uses two natural language inference datasets, MNLI and SNLI,
as out-of-domain source datasets for Coarse-tuning, and two MCQA datasets, DREAM and RACE, as
in-domain source datasets for Multi-task learning. The characteristics of different data sets make the
model have better generalization ability\citep{jin2020mmm}. Third, because these data sets are highly related to ReCAM, it is suitable to use MMM encoder as initial encoder of ComSR.
\begin{equation}\label{eq2}
	\mathbf{H}^{p}={Bert}(\mathbf{p}), \quad \mathbf{H}^{s}={Bert}(\mathbf{s})
\end{equation}

where ${Bert}(\cdot)$ indicate BERT model which return the last attention layer. $\mathbf{H}^{p} \in R^{l \times|p|}, \mathbf{H}^{s} \in R^{l \times|s|}$ are
sequence representation of the passage and summary. $|p|,|s|$ are the length of the passage and summary. $l$
is the dimension of the hidden state.

\subsection{Answer Prediction}
In the answer prediction module, we pass the output of the encoder through a fully connected layer and a
pooling layer to finally get the score of each option.
\begin{equation}\label{eq3}
	s_{k}=\mathbf{v}_{s}^{T} \tanh \left(\mathbf{W}_{\mathbf{k}}\left[\mathbf{H}^{p} ; \mathbf{H}^{s}\right]+\mathbf{b}_{\mathbf{k}}\right)
\end{equation}
where $s_{k}$ is score of the answer, $\mathbf{v}_{s}^{T}$ and $\mathbf{W}_\mathbf{k}$ are learnable parameters. $\tanh (\cdot)$ performs tanh activation function, $\mathbf{W}_{k} \in \mathbf{R}^{l \times l}, \mathbf{b}_{k} \in$
$\mathbf{R}^{l}, \mathbf{v}_{s}^{T} \in \mathbf{R}^{l}$.

The probability $P(k \mid P, S)$ of summary $S_{k}$ to be the correct answer is computed as
\begin{equation}\label{eq4}
	P(k \mid P, S)=\frac{\exp \left(s_{k}\right)}{\sum_{i}^{N=5} \exp \left(s_{i}\right)}
\end{equation}

Then according to $P(k \mid P, S)$, the loss function is defined as
\begin{equation}\label{eq5}
	J(\theta)=-\log (P(k \mid P, S))
\end{equation}

\section{Experimental setup}
\subsection{Dataset}
The data set of Reading Comprehension of Abstract Meaning (ReCAM) is provided by the organizer
of the competition containing a large number of abstract words in answers. Abstract words refer to
thoughts and concepts that are far from immediate perception. In ReCAM, we divide abstract words into
Imperceptibility words and Nonspecificity words, and provide two data sets, respectively. The accuracy
we showed in the paper was tested on the dev dataset.
\subsubsection{ReCAM-Imperceptibility}
Imperceptible words are the words of being Imperceptibility by eyes or senses, such as experience, success,
significant, challenge. This data set provides a passage, a question with a placeholder and five answers
with an imperceptibility word. The training set/test set/validation set contains 3227/2025/837 pieces of
data respectively.

\subsubsection{ReCAM-Nonspecificity}
Nonspecific words are words with very broad concepts, and hypernyms are often words of this type, such
as food, jewelry, people, and vehicle. This data set provides a passage, a question with a placeholder and
five answers with a Nonspecificity word. The training set/test set/validation set contains 3318/2017/851
pieces of data respectively.
\subsection{Implementation Details}
In the test method, we use the accuracy of the model in the answer option of the data set as the measurement
standard. For model training, we will train each tested model for 10 epochs at a learning rate of 3e-05,
use cross-entropy to calculate loss, and use Adam optimizer\citep{kingman2015adam} for fine-tuning. On
hardware devices, we use GeForce RTX-2080Ti to provide computing.

\section{Results}
\subsection{Comparison with Baselines}
Since our model uses BERT as encoder, we use the original BERT\citep{devlin2018bert} for comparison. As pre-trained language models like BERT, we use ALBERT\citep{lan2019albert} and RoBERTa\citep{liu2019roberta} in the experiment. In addition, we also adopt MMM\citep{jin2020mmm} that performed
well in multi-choice data sets such as RACE\citep{lai2017race} and DREAM\citep{sun2019dream} to do experiments.

\begin{table}[h!]
	\begin{center}
		\begin{tabular}{p{0.55\linewidth}p{0.14\linewidth}p{0.14\linewidth}}
			\hline
			{\bf Model}&{\bf Subtask 1}(\%)&{\bf Subtask 2}(\%)\\
			\hline
			BERT-base&47.19&50.65\\
			RoBERTa&22.34&22.44\\
			ALBERT&36.91&37.25\\
			MMM+BERT-base&57.11&59.11\\
			ComSR+BERT-base(ours)&{\bf 64.76}&{\bf 64.86}\\
			\hline
		\end{tabular}
	\end{center}
	\caption{Experimental results on ReCAM. The overall best results are in bold face.\label{tab1}}
\end{table}

\begin{table*}[h!]
	\begin{center}
		\begin{tabular}{p{0.45\textwidth}p{0.15\textwidth}p{0.15\textwidth}}
			\hline
			{\bf Model}&{\bf I$\rightarrow$N} (\%)&{\bf N$\rightarrow$I} (\%)\\
			\hline
			ComSR(Passage + summary)(ours)&50.65(14.11 $\downarrow $)&51.73(13.13 $\downarrow $)\\
			ComSR(Passage + summary + question)&48.53(17.18 $\downarrow $)&49.94(15.51 $\downarrow $)\\
			ComSR(Passage + summary + answer)&44.89(18.55 $\downarrow $)&46.71(18.38 $\downarrow $)\\
			MMM(Passage + question + answer)&39.48(17.63 $\downarrow $)&43.73(15.38 $\downarrow $)\\
			\hline
		\end{tabular}
	\end{center}
	\caption{\label{font-table} Comparison generalization on ReCAM. I$\rightarrow$N denotes training on the Imperceptibility dataset and testing on the Nonspecificity dataset, N$\rightarrow$I is the opposite. $\downarrow $ represents the drop for a model compared to the test in its own test set and + denotes concatenation.\label{tab2}}
\end{table*}

Table~\ref{tab1} show that the model has similar features in task1 and task2. In the experimental
results of task1 and task2, we can find that (1) Although pre-trained language models have slightly different
results depending on the models, the overall level is low on the small data set. (2)
MMM proposed for the MCQA task performs better than the pre-trained language model on ReCAM.
This is because the encoder of MMM uses the Multi-stage Multi-task Learning to train on NLI data sets
such as MNLI and SNLI and on MCQA data sets such as DREAM and RACE, which is more suitable for
ReCAM. (3) The performance of ComSR is better than other baselines, and it has significantly improved.
It can be seen that, under the condition of using the same encoder, replacing the placeholder of question with answer is more conducive to understand sentence meaning.



\subsection{Analysis Studies}
\subsubsection{The completeness of the semantic expression of summary}
Converting question and answer into summary input improves semantic integrity and the
model's ability to understand sentences. However, is the semantics provided by summary complete? To
study this problem, we set up the following experiment.
\begin{table}[h!]
	\begin{center}
		\begin{tabular}{p{0.4\linewidth}p{0.2\linewidth}p{0.2\linewidth}}
			\hline
			{\bf Model}&{\bf Subtask 1}(\%)&{\bf Subtask 2}(\%)\\
			\hline
			ComSR(Passage + summary)(ours)&64.76&64.86\\
			ComSR(Passage + summary + question)&65.71&65.45\\
			ComSR(Passage + summary + answer)&63.44&65.09\\
			\hline
		\end{tabular}
	\end{center}
	\caption{Comparison among different implementation of the input method on ReCAM. + denotes concatenation.\label{tab3} }
\end{table}

If there is a loss of semantics in the synthesis of the summary, then the summary can be supplemented
when the original question and answer are used, thereby improving the performance of the
model. However, it can be seen from the table~\ref{tab3} results that some of the experimental results of the passage+
summary+answer and passage+summary+question groups are equal to passage+summry and even
slightly lower than passage+summary. Therefore, the conversion of question and answer into summary
makes the semantics of options more complete, and there is no semantic loss.

\subsubsection{Research on model generalization}
To study the generalization ability of the model, (1) we use the ReCAM-Imperceptibility data set
to train and test it in the ReCAM-Nonspecificity data set. (2) we use the ReCAM-Nonspecificity data set
to train and test it in the ReCAM-Imperceptibility data set. In order to study the most suitable input form, we use passage+summary+question, passage+summary+answer and MMM (passage+question+answer) model for comparison.

According to the experimental results in table~\ref{tab2}, after switching the data set for testing, the highest
accuracy is ComSR(Passage + summary). Although the results of the test in another data set have dropped, the ComSR using the Passage and summary received the least impact. In summary,
ComSR(Passage + summary) performs best in the above models, and the method of matching summary
and Passage alone has stronger generalization ability.

\subsubsection{Case Analysis}
Table~\ref{tab4} shows a passage and its corresponding question and answers. To understand ComSR's ability to disambiguate, we specifically selected samples that contained polysemous words in answers. In this example,  the answer "goods" is a typical polysemous word.

\begin{table}[h!]
	\begin{center}
		\begin{tabular}{|p{0.25\linewidth}|p{0.65\linewidth}|}
			\hline
			{\bf Passage}& Media playback is unsupported on your device 19 November 2013 Last updated at 19:46 GMT Following the deaths of six cyclists in a period of 14 days, there have been increased calls for restrictions on HGVs during peak times. Paul Hutchin shows reporter Marc Ashdown what it is like driving a lorry in the capital.\\
			\hline
			{\bf Question}&Heavy {\bf @placeholder} vehicles have been involved in nine of this year 's 14 cyclist crash fatalities in London.\\\hline
			{\bf Answers}&(a) operators  {\bf (b) goods}  (c) groups  (d) patrol  (e) air\\
			\hline
		\end{tabular}
		\caption{Example of ReCAM.\label{tab4} }
	\end{center}
\end{table}

\begin{figure}[h!]
	\centering
	\subfigure[MMM(passage + question + answer)]{\includegraphics[width=6.5cm]{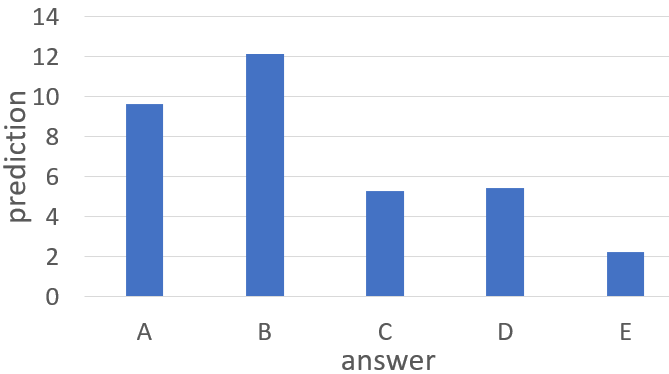}}
	\subfigure[ComSR(passage + summary)]{\includegraphics[width=6.5cm]{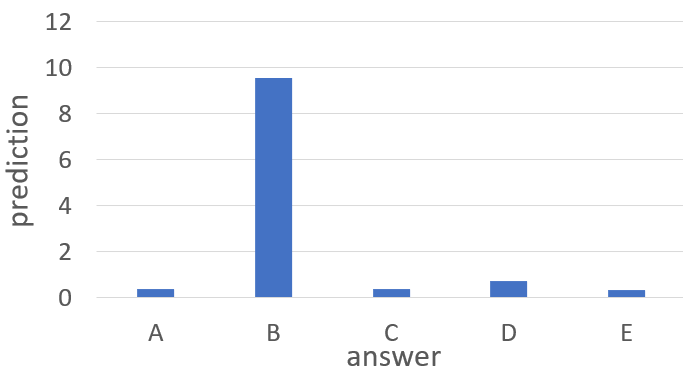}}
	\caption{Prediction distributions in different models. The model selects the answer with the highest predicted value as the output option. \label{fig3} }
\end{figure}

Models take the logarithm of the value to get the predictions. Because many of the predictions are negative, we add bias = 12.5 to the predictions for easy viewing and show it in Figure~\ref{fig3}.

We observe that although MMM selects the right option, the difference between the correct answer and other option values in the probability distribution is not obvious. On the contrary, correct prediction value is much higher than other values in ComSR. By combining question and answer into summary, ComSR can not only improve the semantic integrity of the sentence, but also eliminate the ambiguity caused by the polysemy of the original answer.
Therefore, even for polysemous words that are prone to ambiguity, ComSR can obtain accurate answers by fully understanding the summary representation.


\section{Conclusion}
We proposed ComSR for multiple choice reading comprehension to complete summary representation. Merging the separated question and answer into a summary could improve semantic integrity and better reveal the relationship between option and question.
Results showed that our model can achieve relatively high performance compared to the latest baseline in
terms of accuracy and generalization ability.

\bibliographystyle{acl_natbib}
\bibliography{acl2021}


\end{document}